\documentclass[runningheads]{llncs}

\usepackage[T1]{fontenc}
\usepackage{microtype}
\usepackage{graphicx}
\usepackage{amsmath}
\usepackage{amssymb}
\usepackage{textcomp}
\usepackage{booktabs}
\usepackage{float}
\usepackage{placeins}
\usepackage{url}
\usepackage{xcolor}
\usepackage{tabularx}
\usepackage{multirow}
\usepackage{tikz}
\usepackage{booktabs} 
\usepackage{multicol}
\usepackage{subfigure}
\usepackage{threeparttable}
\usepackage{mathrsfs}
\usepackage{makecell}
\usepackage{hyperref}

\makeatletter
\newcommand{\thickhline}{%
\noalign {\ifnum 0=`}\fi \hrule height 1pt
\futurelet \reserved@a \@xhline
}
\makeatother

\usepackage[table]{xcolor}       
\definecolor{mygray}{gray}{.9}  
\PassOptionsToPackage{numbers}{natbib}
\definecolor{url}{RGB}{0,73,147}
\definecolor{mypink}{HTML}{bc4749}

\usepackage{xspace}
\usepackage{bbding}
\usepackage{wrapfig}
\makeatletter
\DeclareRobustCommand\onedot{\futurelet\@let@token\@onedot}
\def\@onedot{\ifx\@let@token.\else.\null\fi\xspace}

\makeatletter
\DeclareRobustCommand\onedot{\futurelet\@let@token\@onedot}
\def\@onedot{\ifx\@let@token.\else.\null\fi\xspace}

\def\etal{\emph{et al}\onedot}
\makeatother

\begin{document}

\title{Motion Reinforces Appearance: RGB-Skeleton Gated Residual Fusion for Micro-Gesture Online Recognition}
\titlerunning{RGB-Skeleton Fusion for Micro-Gesture Recognition}

\author{Jialin Liu \and Xinwen He \and Pengyu Liu \and Jiale Shi \and Huaijuan Zang \and \\  Yanbin Hao$^{\dagger}$ }

\authorrunning{J. Liu \etal} 

\institute{School of Computer Science and Information Engineering, Hefei University of Technology (HFUT), Hefei, China \\
\email{lin1557173497@gmail.com, hxw66601@gmail.com, lpynow@gmail.com, shijiale2006@hotmail.com, zanghj@hfut.edu.cn, haoyanbin@hotmail.com}} 

\footnotetext[1]{$^{\dagger}$: Corresponding author.}

\maketitle

\begin{abstract}
\begingroup
\emergencystretch=2.5em
\hyphenpenalty=2000
\exhyphenpenalty=2000
Micro-gesture analysis attracts increasing attention for inferring spontaneous emotion from subtle body movements. Micro-gesture online recognition, which localizes and classifies each gesture instance in untrimmed videos, is a core task in the 4th EI-MiGA-IJCAI Challenge. 
Compared with typical temporal action detection, MGR emphasizes the localization and classification of actions, requiring the model to output the start time, end time, and category of each micro-gesture. Moreover, since micro-gestures are highly spontaneous, relying solely on a single modality makes it difficult to capture the complete and accurate multi-modal cues.
In this work, we propose DyFADet+, which extends DyFADet into a dual-stream RGB-skeleton framework. In our model, both modalities are projected into shared multi-scale temporal embeddings and fused through a gated residual module, which adaptively injects skeleton motion into the RGB representation rather than using naive concatenation. Finally these fused features are decoded by a Dynamic TAD head
for online classification and boundary regression. On the SMG dataset, our method achieves an $F_1$ score of 40.88, ranking 2nd in the Micro-gesture Online Recognition track.

\endgroup

\keywords{Micro-gesture online recognition, micro action, video understanding}
\end{abstract}

\section{Introduction}

Body gestures are an important form of non-verbal communication and can reveal emotional states when facial expressions or speech are deliberately controlled. Subtle body movements are often harder to suppress and thus useful for inferring hidden affect~\cite{calvo2010affect,imigue2021cvpr}. Micro-gestures are brief, spontaneous movements, such as scratching the head or touching the nose, that arise under stress or discomfort, unlike intentional communicative gestures that accompany speech~\cite{smg2023ijcv,hao2022attention,hao2022group,imigue2021cvpr}. 
Understanding such cues supports applications in affective computing, human-computer interaction~\cite{Noroozietal2021}, and security-sensitive interview analysis~\cite{Poppeetal2024}, where the goal is to infer spontaneous or suppressed emotion from behavior rather than from explicit verbal reports~\cite{Ekman1969,liu2025survey}.

Micro-gesture Online Recognition~\cite{miga2challenge2024,miga2024} requires methods to parse each untrimmed video into a set of gesture instances. For every detected segment, the method must predict both temporal boundaries and one of 17 fine-grained categories on the spontaneous Micro-Gesture dataset~\cite{smg2023ijcv}. Unlike offline clip-level classification, MGR requires both temporal localization and fine-grained categorization within continuous, untrimmed videos. 
Most gesture and action recognition research still targets macro actions~\cite{posec3d2022,li2023data,stgcn2018}, and general temporal action detection methods are typically designed for longer, more salient motions with clearer boundaries~\cite{dyfadet2024,actionformer2022}. 
Unlike macro actions, MGR faces unique challenges due to low-amplitude, weakly salient motions with fuzzy boundaries. Furthermore, micro-gestures are frequently obscured by irrelevant body movements~\cite{smg2023ijcv,shang2025cross}.
While RGB streams retain rich appearance details, they often lack reliable motion cues for such fine-grained dynamics. Conversely, skeleton data provide distinct pose kinematics, but can be highly noisy due to the small displacements of micro-gestures~\cite{wang2024miga_dualstream}. Therefore, how to fully harness these two modalities to mitigate their respective weaknesses has become a critical challenge in MGR.
Prior MGR methods explore dedicated skeleton-based graph convolution models~\cite{guo2023miga_online}, dual-stream Transformers~\cite{wang2024miga_dualstream}, or query-based temporal designs~\cite{liu2024miga_query}. Meanwhile, recent micro-gesture recognition methods investigate skeletal embeddings~\cite{li2023joint}, prototype learning~\cite{chen2024prototype,li2025prototypical}, and multimodal fusion~\cite{gu2025mm}. However, how to fuse RGB and skeleton inside a unified temporal detector remains underexplored. In particular, naive early fusion or concatenation may amplify pose-estimation errors instead of exploiting cross-modal complementarity.

To address these issues, we extend DyFADet~\cite{dyfadet2024} into a dual-stream RGB-Skeleton architecture for MGR. Under this framework, RGB and skeleton inputs are first projected to a shared multi-scale temporal representation. In parallel, the RGB branch follows DyFADet’s dynamic embedding projection, while the skeleton branch uses a convolutional Transformer projector with local temporal attention. At the fusion stage, a gated residual fusion module adaptively injects skeleton motion into the RGB features at each pyramid level. In particular, a lightweight adapter maps skeleton cues to the RGB channel space, and a sigmoid gate conditioned on both modalities decides where to trust kinematic information. 
Ultimately, the fused pyramid is decoded by the detection head for online classification and boundary regression. This design preserves discriminative appearance while selectively reinforcing motion-sensitive regions, effectively avoiding the instability of unconstrained concatenation.

The main contributions of this work are summarized as follows:
\begin{itemize}
    \item We propose DyFADet+, a dual-stream RGB-Skeleton framework tailored for online micro-gesture recognition. By projecting both modalities into shared multi-scale temporal embeddings, our model effectively captures the short and subtle dynamics of micro-gestures.
    \item We design an innovative RGB-Skeleton Gated Residual Fusion module to leverage multi-modal cues. Instead of relying on naive concatenation, this module adaptively injects fine-grained skeleton motion into the RGB representation, providing precise features for the Dynamic TAD head to perform robust online classification and boundary regression.
  \item We validate our approach on the MiGA challenge benchmark\protect\footnotemark[1], achieving $F_1{=}40.88$ and 2nd place on the leaderboard. Meanwhile, Extensive experiments on the SMG benchmark show that our method effectively enhances micro-gesture detection.
\end{itemize}

\footnotetext[1]{The Kaggle competition page: \href{https://www.kaggle.com/competitions/the-4th-ei-mi-ga-ijcai-challenge-track-2/leaderboard}{https://www.kaggle.com/competitions/the-4th-ei-mi-ga-ijcai-challenge-track-2/leaderboard}}

\section{Related Work}

\subsection{Micro-Gesture Datasets}

Research on subtle body behaviour was driven by diverse benchmarks that differed in collection setting, annotation scales, and evaluation metrics~\cite{gu2025motion,ma522024tcsvt,li2026bench,mmad2025iccv}. For instance, iMiGUE~\cite{imigue2021cvpr} was a video dataset collected from many subjects and cultural backgrounds under realistic interview conditions. It provided dense frame-level micro-gesture annotations so that models had to infer suppressed or hidden affect from non-verbal cues rather than from subject identity or staged performance. SMG~\cite{smg2023ijcv} focused on naturally occurring micro-gestures induced under psychological stress. It offered multimodal inputs including RGB and skeleton data, together with analyses linking gesture patterns to emotional stress. It defined 17 fine-grained categories and 40 samples. MA-52~\cite{ma522024tcsvt} broadened the classification scheme to 52 whole-body micro-action categories with 7 body-part labels. These data were captured from face-to-face psychological interviews involving 205 participants and contained 22,422 trimmed clips for clip-level recognition with whole-body coverage. MMA-52~\cite{mmad2025iccv} further annotated untrimmed videos with start and end times and multiple co-occurring labels per interval, which formulated multi-label micro-action detection closer to temporal action detection than clip-level classification. 

\subsection{Temporal Action Detection}

Micro-gesture online recognition coupled fine-grained classification with temporal localization in untrimmed videos~\cite{miga2challenge2024,smg2023ijcv}. Due to the subtle amplitude and brief duration of micro-gestures, their temporal boundaries remained inherently ambiguous. Furthermore, background body movements often co-occurred with target actions, which aligned the task closely with temporal action detection to predict precise intervals and category labels from continuous features. 
Existing methods handled these difficulties by capturing fine-grained structural and temporal context.
Guo \emph{et al.}~\cite{guo2023miga_online} combined graph convolutions and multiscale Transformers to capture both joint structures and long-range dependencies. Wang \emph{et al.}~\cite{wang2024miga_dualstream} proposed a dual-stream multiscale Transformer that processed RGB appearance and skeleton motion in parallel and fused multi-resolution temporal features for long interview videos. 
Liu \emph{et al.}~\cite{liu2024miga_query} introduced learnable query points that attend to informative temporal locations before classification and localisation, and further explored data augmentation with spatial-temporal attention for online recognition~\cite{liu2025online}.
Yet, these methods emphasized specialized, modality-specific encoders rather than integrating strong pre-extracted RGB and skeleton descriptors inside a mature TAD decoder.

In parallel, general TAD methods that matured on benchmarks such as THUMOS-14~\cite{thumos14} and ActivityNet~\cite{activitynet2015} became available in the OpenTAD toolbox~\cite{opentad2025}. Early milestones primarily relied on two-stage paradigms. For instance, BMN~\cite{bmn2019} generated complete candidate intervals by learning boundary-matching maps, while G-TAD~\cite{gtad2020} localized actions by matching sub-graphs constructed over video snippets. However, to bypass the complex and computationally expensive proposal generation process, the field subsequently shifted toward point-based architectures. Pioneering this shift, AFSD~\cite{afsd2021} exploited salient boundary cues for precise extent prediction without predefined reference windows. Building upon this, ActionFormer~\cite{actionformer2022} significantly advanced the field by stacking transformer encoders with an FPN-style head to regress boundaries at multiple resolutions, and TriDet~\cite{tridet2023} further refined boundary quality via three-branch relative distance modeling. Despite these advances, managing complex action overlaps and high computational overhead remained challenging.To address these bottlenecks, PointTAD~\cite{pointtad2022} adopted learnable query points for joint multi-label classification and localization. More recently, DyFADet~\cite{dyfadet2024} introduced dynamic feature aggregation across pyramid levels with a dynamic embedding projection and a DyHead decoder, successfully striking an optimal balance between detection performance and computational efficiency. 
Since direct concatenation may corrupt RGB features with noisy skeleton data, we avoid this naive method. Instead, we propose DyFADet+, which extends the original DyFADet into a dual-stream framework using a gated residual fusion module. This module adaptively injects skeleton motion cues into the RGB feature pyramid before the detection head. In this way, our approach keeps the RGB representation stable, allowing the network to automatically learn when and where the skeleton data is reliable.

\subsection{Skeleton-Based Action Recognition}

Skeleton sequences provide a clear way to represent joint movements without background interference. This makes them highly effective for micro-gesture analysis. Even when subtle hand or arm motions are obscured by cluttered backgrounds in RGB videos, they can still be accurately captured by the skeleton data.Early graph-based recognizers modeled the human skeleton as a spatio-temporal graph. Specifically, ST-GCN~\cite{stgcn2018} applied graph convolutions with explicit spatial partitioning over joints and temporal neighborhoods. Subsequently, 2s-AGCN~\cite{2sagcn2019} learned data-driven adjacency for joint and bone streams to capture action-specific structures. Furthermore, CTR-GCN~\cite{ctrgcn2021} refined connectivity channel-wise so that different body parts attended to distinct relational patterns. Alternatively, PoseC3D~\cite{posec3d2022} stacked joint heatmaps into 3D volumes processed by 3D CNNs. This approach combined convolutional locality with explicit temporal stacking, which achieved strong results on large-scale action benchmarks.

For micro-gesture online recognition, MiGA solutions typically embedded skeletons inside end-to-end temporal architectures, using either graph convolutions coupled with multiscale Transformers~\cite{guo2023miga_online} or parallel RGB-skeleton Transformer encoders~\cite{wang2024miga_dualstream}. This ensured that joint structures and long-range temporal context were learned jointly with localization. Furthermore, data augmentation on pose sequences was shown to improve behavior analysis in multi-person settings~\cite{li2023data}. However, such designs depended on pose estimates extracted from video and could be highly sensitive to estimation noise when movements were low-amplitude. In contrast, we treated the skeleton as a collaborative motion stream built from pretrained embeddings, projected them to the same multi-scale temporal pyramid as the RGB appearance, and integrated both through gated residual fusion before DyFADet decoding. This ensured that kinematic cues were selectively reinforced, rather than replaced, discriminative RGB features when pose information was reliable.
\section{Method}\label{sec:method}

\subsection{Task Definition}


We formulate micro-gesture online recognition as a temporal action detection problem on the SMG dataset~\cite{miga2challenge2024,smg2023ijcv}. Given an untrimmed video sequence $\mathcal{V}$, we represent it through a dual-stream feature representation to capture both appearance and kinematics. Specifically, the video is encoded into an RGB feature sequence $\mathbf{V}^{r} = \{\mathbf{v}^{r}_t\}_{t=1}^T \in \mathbb{R}^{T \times d_{r}}$ using pre-trained visual encoders (\emph{e.g.}, VideoMAE~\cite{videomae2022}), alongside a temporally aligned skeleton feature sequence $\mathbf{V}^{s} = \{\mathbf{v}^{s}_t\}_{t=1}^T \in \mathbb{R}^{T \times d_{s}}$. 
The objective is to predict a set of action instances $\Psi = \{\psi_1, \psi_2, \ldots, \psi_N\}$, where $N$ denotes the total number of predicted instances. Each instance $\psi_i = (t_{i}^{s}, t_{i}^{e}, c_i)$ is characterized by its start time $t_{i}^{s}$, end time $t_{i}^{e}$, and category label $c_i \in \mathcal{C}$, where $\mathcal{C}$ is the set of all predefined micro-gesture categories.

\subsection{Pipeline}
\begin{figure}[t]
\centering
\resizebox{\linewidth}{!}{\rotatebox{-90}{\includegraphics{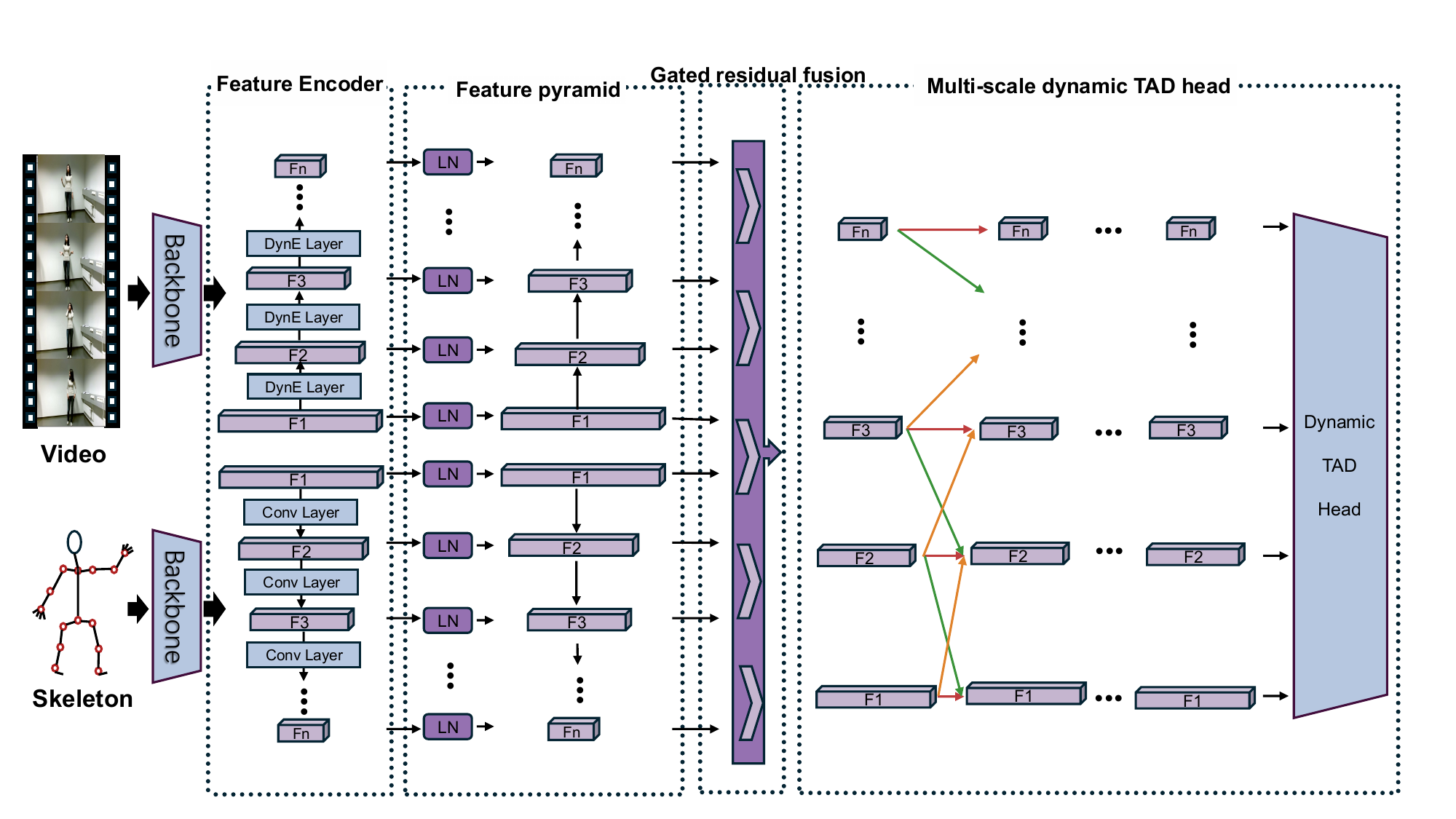}}}
\caption{Overview of the proposed DyFADet+ architecture for SMG micro-gesture detection.}
\label{fig:framework}
\end{figure}

We build DyFADet+ on DyFADet~\cite{dyfadet2024}, a dynamic feature aggregation detector for temporal action localization, and replace its single-stream projection with a two-stream RGB-Skeleton design. RGB appearance and skeleton motion are encoded into a shared multi-scale representation, fused with a gated residual module, normalized with the same lightweight multi-scale neck, and decoded by the dynamic detection head. The overall pipeline is illustrated in Fig.~\ref{fig:framework}.

\subsection{RGB and Skeleton Feature}


The RGB branch adopts the projection architecture of DyFADet~\cite{dyfadet2024}, employing a shallow temporal convolutional embedder followed by a sequence of Dynamic Encoder (DynE) blocks. By leveraging dynamic feature aggregation and progressive temporal pooling, these blocks systematically downsample the sequence to construct a multi-scale feature pyramid $\{\mathbf{F}_{r}^{(\ell)}\}_{\ell=1}^{L}$.
Parallel to the DynE blocks in the RGB branch, the skeleton branch employs a lightweight architecture driven by Conv blocks. We first embed the raw skeletal features into a higher-dimensional space using 1D convolutions. This is followed by initial Conv blocks that refine the representation at the original temporal resolution. A subsequent series of Conv blocks progressively downsamples the temporal dimension to construct a corresponding multi-scale pyramid $\{\mathbf{F}_{s}^{(\ell)}\}_{\ell=1}^{L}$ that is structurally aligned with the RGB branch. To guarantee temporal alignment prior to fusion, we apply nearest-neighbor interpolation to upsample the skeleton features whenever their temporal length is shorter than the corresponding RGB counterpart.

\begin{figure}[t]
\centering
\resizebox{\linewidth}{!}{\rotatebox{-90}{\includegraphics{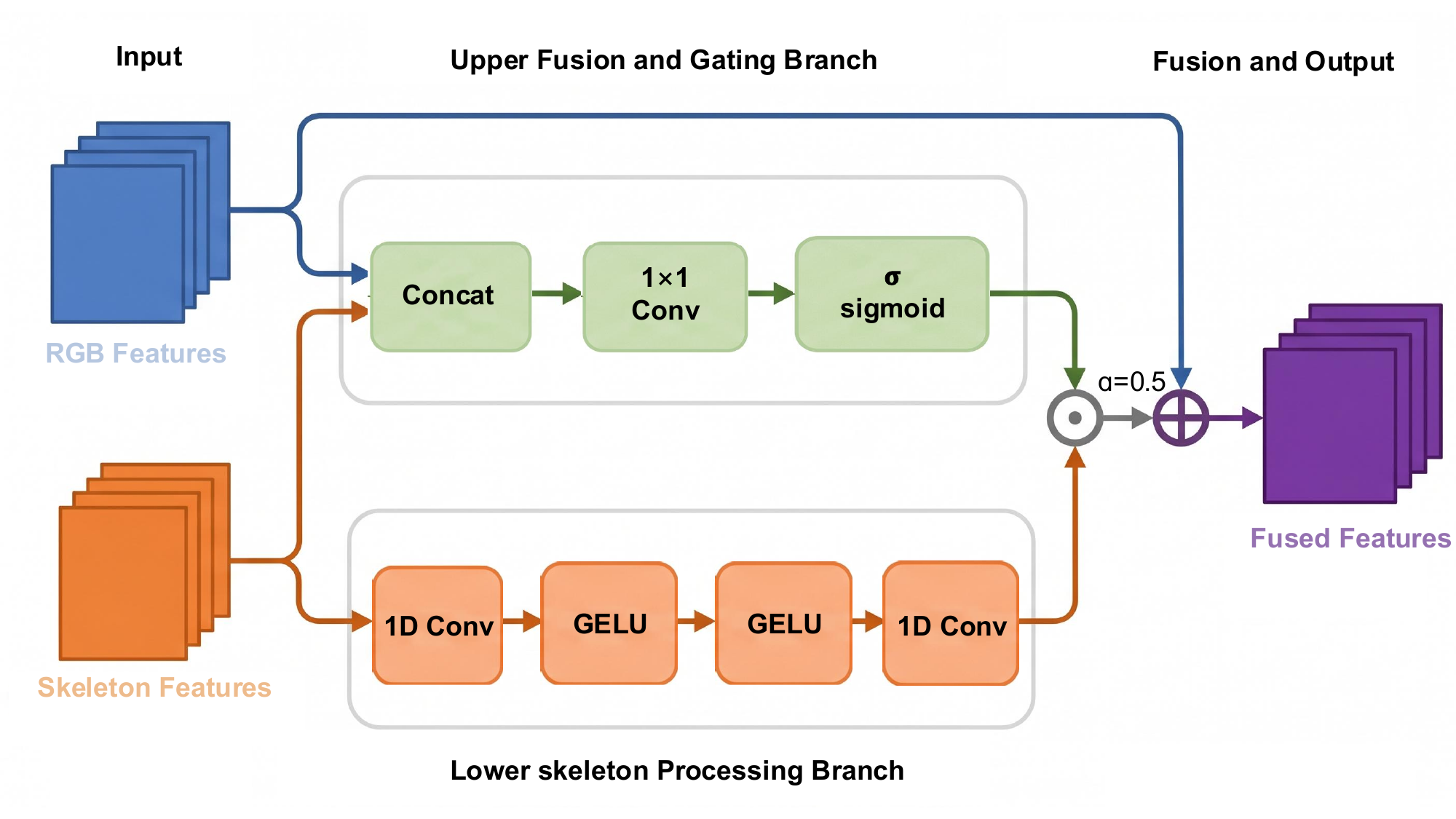}}}
\caption{Gated residual fusion at a single pyramid level: skeleton features pass through an adapter, a sigmoid gate (conditioned on RGB and skeleton) selects motion cues, and the result is added to RGB with scaling $\alpha$.}
\label{fig:fusion}
\end{figure}

\subsection{Gated Residual Fusion}

Naive early fusion or concatenation can amplify noise when pose estimation is unstable. We instead inject skeleton cues as a gated residual on top of the RGB features at every pyramid level $\ell$. A lightweight adapter $\psi_\ell(\cdot)$ maps the skeleton features to the RGB channel space, while a level-specific gate observes both modalities:
\begin{align}
\mathbf{G}^{(\ell)} &= \sigma\!\left(\phi_\ell\!\left(\left[\mathbf{F}_{r}^{(\ell)},\mathbf{F}_{s}^{(\ell)}\right]\right)\right),\label{eq:fusion_gate}\\
\mathbf{F}_{f}^{(\ell)} &= \mathbf{F}_{r}^{(\ell)} + \alpha\cdot \mathbf{G}^{(\ell)} \odot \psi_\ell\!\left(\mathbf{F}_{s}^{(\ell)}\right),\label{eq:fusion_fused}
\end{align}
where the bracketed term $[\mathbf{F}_{r}^{(\ell)},\mathbf{F}_{s}^{(\ell)}]$ concatenates the RGB and skeleton features along the channel dimension (the per-time-step feature axis); $\phi_\ell$ is implemented as a $1{\times}1$ temporal convolution, $\sigma$ is the sigmoid function, $\odot$ denotes element-wise multiplication, and $\alpha{=}0.5$ scales the residual branch. The adapter $\psi_\ell$ is a two-layer temporal convolutional stack with GELU activations. Each level ends with temporal layer normalization and dropout ($p{=}0.05$) on the skeleton residual. We initialize the bias of the gate convolution to $-2$ so that $\mathbf{G}^{(\ell)}$ starts near zero. 
This stabilizes early training. It forces the network to focus on the RGB path first, preventing noisy skeleton features from harming performance.

\subsection{Temporal Detection Head and Losses}

Upon obtaining the fused multi-scale pyramid $\{\mathbf{F}_{f}^{(\ell)}\}_{\ell=1}^{L}$, the features are directly fed into the dynamic detection head~\cite{dyfadet2024}. Following the mechanism of DyFADet, the DyHead dynamically adjusts the shared parameters based on the input and selectively fuses cross-level features. The final predictions are yielded by two parallel branches across all levels, namely a classification module and a regression module. The classification module employs a 1D convolution followed by a Sigmoid function to predict the probabilities of the K=17 micro-gesture categories at each timestamp. Concurrently, the Reg module utilizes a 1D convolution with a ReLU activation to estimate the distances from the current timestamp to the start and end boundaries of the action instances. To handle micro-gestures of varying durations, we generate dense temporal points using predefined strides of $\{1,2,4,8,16,32\}$, with corresponding scale-aware regression ranges of $(0,4)$, $(4,8)$, $(8,16)$, $(16,32)$, $(32,64)$, and $(64,\infty)$. The entire framework is optimized end-to-end using the focal loss~\cite{lin2017focal} for classification and the Distance-IoU (DIoU) loss~\cite{zheng2020distanceiou} for segment regression.
\section{Experiments}\label{sec:experiments}

\subsection{Dataset and Evaluation Metric}

\noindent\textbf{Dataset.}
We conduct experiments on the SMG dataset~\cite{smg2023ijcv}, which contains 3,692 annotated micro-gesture instances across 17 categories collected from 40 subjects. Crucially, the dataset provides RGB videos and skeleton sequences, making it an ideal benchmark for our proposed dual-stream architecture. The 40 subjects are divided into non-overlapping splits of 30 for training, 5 for validation, and 5 for testing.

\noindent\textbf{Evaluation Metric.}
We jointly assess detection and classification using the $F_1$ score:
\begin{equation}
F_1 = \frac{2 \cdot \mathrm{Precision} \cdot \mathrm{Recall}}{\mathrm{Precision} + \mathrm{Recall}}.
\end{equation}
For a long evaluation sequence, Precision is the fraction of correctly classified micro-gestures among all instances predicted by the model in that sequence, and Recall is the fraction of ground-truth micro-gestures that are both detected and correctly labeled.
This metric comprehensively reflects the ability of a method to both detect and correctly classify micro-gestures.

\subsection{Implementation Details}
RGB features are extracted using VideoMAEv2-g~\cite{videomae2022}, while the skeleton inputs utilize pre-computed kinematic representations.
During training, we apply random temporal truncation. The feature stride is set to 4, and the sample stride to 1. The model is trained with AdamW~\cite{loshchilov2017decoupled} using an initial learning rate of 1e-4 and a weight decay of 0.025. We employ a cosine annealing learning rate schedule with a linear warm-up of 20 epochs, training for a total of 200 epochs. The training batch size is 2, and the validation and testing batch sizes are 1. Gradient clipping with a maximum norm of 1 and an exponential moving average (EMA) are applied.

\subsection{Experimental Results}

\begingroup
\setlength{\intextsep}{8pt plus 2pt minus 2pt}%
\begin{table}[H]
\centering 
\caption{Leaderboard of the top submissions of the past four MiGA Track-2 editions (2023--2026) on the SMG online recognition track. Modalities: R=RGB, S=Skeleton.}
\label{tab:leaderboard}
\renewcommand{\arraystretch}{1}%
\resizebox{\linewidth}{!}{%
\begin{tabular}{|c|c|c|c|c|} 
\hline
\rowcolor{gray!20} Rank & Team & Core Methodology & Modality & $F_1$ \\
\hline
MiGA'23 1st & NPU-Stanford \cite{guo2023miga_online} & Graph-Conv+Multiscale Transformer & S & 14.81 \\
MiGA'23 2nd & HFUT-VUT  & ST-GCN   & S  & 4.67 \\
\hline
MiGA'24 1st & NPU-MUCIS \cite{wang2024miga_dualstream} & PoseConv3D & R + S & 27.57 \\
MiGA'24 2nd & HFUT-VUT \cite{liu2024miga_query} &  I3D+Mamba & R & 14.35 \\
\hline
MiGA'25 1st & HFUT-VUT \cite{liu2025online} & VideoMAEv2-g + DyFADet & R  & 38.03 \\
MiGA'25 2nd & Chutian Meng \cite{meng2025online}  & VideoMAE-g + DyFADet & R & 31.54 \\
\hline
MiGA'26 1st & XInsight Lab \cite{shen2026spatial} & VideoMAEv2-g + AdaTAD & R & 43.81 \\ 

MiGA'26 3rd & XD-L & - & - & 35.56 \\ \hline\hline

\textbf{MiGA'26 2nd} & \textbf{AIM (Ours)} & \textbf{VideoMAEv2-g + DyFADet} & \textbf{R + S} & \textbf{40.88} \\
\hline
\end{tabular}%
}
\end{table}
\endgroup

Table~\ref{tab:leaderboard} summarizes the top-performing submissions across the past four editions of the MiGA challenge from 2023 to 2026 on the SMG online recognition track. The leaderboard highlights a clear evolutionary trajectory in micro-gesture online recognition. Early solutions primarily relied on skeleton-based architectures such as Graph-Conv and ST-GCN. In contrast, recent advancements are heavily driven by powerful foundational video models like VideoMAEv2-g. Notably, the state-of-the-art $F_1$ score has surged from 14.81 in 2023 to over 43.81 in 2026, reflecting the rapid progress in representation learning for fine-grained video understanding. In the 2026 edition, our AIM team secures the second place on the official leaderboard with an $F_1$ score of 40.88. By effectively integrating VideoMAEv2-g appearance features with structural skeleton cues through the DyFADet pipeline, our approach significantly outperforms the 2025 winning solution score of 38.03 by 2.85 points. The first-place submission by XInsight Lab        
~\cite{shen2026spatial} achieves 43.81 using an AdaTAD framework operating solely on RGB inputs. However, our dual-stream design combining RGB and skeleton demonstrates highly competitive performance. This result proves the continued viability and robustness of explicitly fusing joint kinematics with appearance representations for temporal micro-gesture localization.

\subsection{Ablation Study}

\begingroup
\setlength{\intextsep}{8pt plus 2pt minus 2pt}%
\begin{table}[H]
\centering 
\caption{Ablation study of different modality integration strategies on the SMG test set.}
\label{tab:ablation}
\small
\setlength{\tabcolsep}{16pt} 
\renewcommand{\arraystretch}{1.2}%
\begin{tabular}{c l c} 
\toprule
\textbf{ID} & \textbf{Variant } & \textbf{$F_1$} \\
\midrule
1 & RGB only & 38.76 \\
2 & Skeleton only & 22.31 \\
3 & RGB + Skeleton  & 38.96 \\
\midrule
\textbf{4} & \textbf{Ours} & \textbf{40.88} \\
\bottomrule
\end{tabular}
\end{table}
\endgroup
Table~\ref{tab:ablation} shows the ablation results to evaluate different fusion methods. Using only RGB features achieves an $F_1$ score of 38.76, while using only skeleton features gets 22.31. This confirms that RGB appearance provides the main information for micro-gesture recognition. Directly concatenating the two modalities only improves the RGB baseline by 0.20, which means a simple merge cannot effectively use skeleton information. In contrast, our gated residual fusion achieves an $F_1$ score of 40.88. This is 2.12 higher than the RGB baseline and 1.92 higher than direct concatenation. These improvements prove that our learnable gate can successfully select useful skeleton cues and filter out noise, creating a more robust representation for detection.
\section{Conclusion}

In this paper, we tackle the problem of micro-gesture online recognition, where naively fusing multi-modal cues often introduces disruptive noise. To address this, we propose DyFADet+, a novel dual-stream framework, as our solution for the MiGA challenge. Specifically, our method pairs RGB appearance features with skeleton motion descriptors. Rather than using conventional concatenation, DyFADet+ employs a gated residual module for multi-modal integration. This gating mechanism explicitly learns to selectively incorporate reliable skeleton kinematics while suppressing noisy pose estimates, yielding a highly robust multi-scale representation for temporal detection. Evaluated on the SMG benchmark, our method achieves an $F_1$ score of 40.88, securing 2nd place on the official leaderboard and outperforming last year's winning entry. Ablation studies further validate that our learnable cross-modal integration is superior to single-modality baselines and naive fusion. Future work will focus on stronger temporal modeling, long-tail learning, and effective multi-modal alignment to improve generalization on spontaneous micro-gestures.

\section*{Acknowledgments}

This work was supported in part by the National Natural Science Foundation of China (No. 62472393), and by the Industry-University Cooperation Collaborative Education Project of the Ministry of Education (No. 250603873093026).

\bibliographystyle{splncs04}
\bibliography{references}

@inproceedings{opentad2025,
  title={{OpenTAD}: A Unified Framework and Comprehensive Study of Temporal Action Detection},
  author={Liu, Shuming and Zhao, Chen and Zohra, Fatimah and Soldan, Mattia and Pardo, Alejandro and Xu, Mengmeng and Alssum, Lama and Ramazanova, Merey and Alc{\'a}zar, Juan Le{\'o}n and Cioppa, Anthony and Giancola, Silvio and Hinojosa, Carlos and Ghanem, Bernard},
  booktitle={Proceedings of the IEEE/CVF Conference on Computer Vision and Pattern Recognition Workshops},
  pages={2650--2660},
  year={2025}
}

@inproceedings{dyfadet2024,
  title={{DyFADet}: Dynamic Feature Aggregation for Temporal Action Detection},
  author={Yang, Le and Zheng, Ziwei and Han, Yizeng and Cheng, Hao and Song, Shiji and Huang, Gao and Li, Fan},
  booktitle={Proceedings of the European Conference on Computer Vision},
  year={2024}
}

@inproceedings{actionformer2022,
  title={ActionFormer: Localizing Moments of Actions with Transformers},
  author={Zhang, Chen-Lin and Wu, Jianxin and Li, Yin},
  booktitle={Proceedings of the European Conference on Computer Vision},
  pages={492--510},
  year={2022},
  organization={Springer}
}

@inproceedings{bmn2019,
  title={BMN: Boundary-Matching Network for Temporal Action Proposal Generation},
  author={Lin, Tianwei and Liu, Xiao and Li, Xin and Ding, Errui and Wen, Shilei},
  booktitle={Proceedings of the IEEE/CVF International Conference on Computer Vision},
  pages={3889--3898},
  year={2019}
}

@inproceedings{gtad2020,
  title={G-TAD: Sub-Graph Localization for Temporal Action Detection},
  author={Xu, Mengmeng and Zhao, Chen and Rojas, David S. and Thabet, Ali and Ghanem, Bernard},
  booktitle={Proceedings of the IEEE/CVF Conference on Computer Vision and Pattern Recognition},
  pages={10156--10165},
  year={2020}
}

@inproceedings{afsd2021,
  title={Learning Salient Boundary Feature for Anchor-Free Temporal Action Localization},
  author={Lin, Chuming and Li, Jian and Wang, Yabiao and Tai, Ying and Luo, Donghao and Cui, Zhen and Wang, Chengjie and Li, Jilin and Huang, Feiyue and Ji, Rongrong},
  booktitle={Proceedings of the IEEE/CVF Conference on Computer Vision and Pattern Recognition},
  pages={3320--3329},
  year={2021}
}

@inproceedings{tridet2023,
  title={TriDet: Temporal Action Detection with Relative Boundary Modeling},
  author={Shi, Dingfeng and Zhong, Yujie and Cao, Qiong and Ma, Lin and Li, Jia and Tao, Dacheng},
  booktitle={Proceedings of the IEEE/CVF Conference on Computer Vision and Pattern Recognition},
  pages={18857--18866},
  year={2023}
}

@inproceedings{stgcn2018,
  title={Spatial Temporal Graph Convolutional Networks for Skeleton-Based Action Recognition},
  author={Yan, Sijie and Xiong, Yuanjun and Lin, Dahua},
  booktitle={Proceedings of the AAAI Conference on Artificial Intelligence},
  volume={32},
  number={1},
  year={2018}
}

@inproceedings{2sagcn2019,
  title={Two-Stream Adaptive Graph Convolutional Networks for Skeleton-Based Action Recognition},
  author={Shi, Lei and Zhang, Yifan and Cheng, Jian and Lu, Hanqing},
  booktitle={Proceedings of the IEEE/CVF Conference on Computer Vision and Pattern Recognition},
  pages={12026--12035},
  year={2019}
}

@inproceedings{ctrgcn2021,
  title={Channel-Wise Topology Refinement Graph Convolution for Skeleton-Based Action Recognition},
  author={Chen, Yuxin and Zhang, Ziqi and Yuan, Chunfeng and Li, Bing and Deng, Ying and Hu, Weiming},
  booktitle={Proceedings of the IEEE/CVF International Conference on Computer Vision},
  pages={13359--13368},
  year={2021}
}

@inproceedings{posec3d2022,
  title={Revisiting Skeleton-Based Action Recognition},
  author={Duan, Haodong and Zhao, Yue and Chen, Kai and Lin, Dahua and Dai, Bo},
  booktitle={Proceedings of the IEEE/CVF Conference on Computer Vision and Pattern Recognition},
  pages={2969--2978},
  year={2022}
}

@misc{miga2024,
  title={MiGA: Micro-Gesture Analysis for Hidden Emotion Understanding Challenge},
  author={{MiGA Organizers}},
  howpublished={\url{https://cv-ac.github.io/MiGA2/}},
  year={2024},
  note={Accessed: 2026-05-10}
}

@inproceedings{miga2challenge2024,
  author={Chen, Haoyu and Schuller, Bj{\"o}rn W. and Adeli, Ehsan and Zhao, Guoying},
  title={The 2nd Challenge on Micro-gesture Analysis for Hidden Emotion Understanding ({MiGA}) 2024: Dataset and Results},
  booktitle={Proceedings of the IJCAI 2024 Workshop on Micro-gesture Analysis for Hidden Emotion Understanding (MiGA 2024)},
  series={CEUR Workshop Proceedings},
  volume={3848},
  year={2024}
}

@article{ma522024tcsvt,
  title={Benchmarking Micro-action Recognition: Dataset, Methods, and Applications},
  author={Guo, Dan and Li, Kun and Hu, Bin and Zhang, Yan and Wang, Meng},
  journal={IEEE Transactions on Circuits and Systems for Video Technology},
  year={2024},
  note={arXiv:2403.05234}
}

@inproceedings{mmad2025iccv,
  title={{MMAD}: Multi-label Micro-Action Detection in Videos},
  author={Li, Kun and Liu, Pengyu and Guo, Dan and Wang, Fei and Wu, Zhiliang and Fan, Hehe and Wang, Meng},
  booktitle={Proceedings of the IEEE/CVF International Conference on Computer Vision},
  year={2025}
}

@inproceedings{imigue2021cvpr,
  title={{iMiGUE}: An Identity-Free Video Dataset for Micro-Gesture Understanding and Emotion Analysis},
  author={Liu, Xin and Shi, Henglin and Chen, Haoyu and Yu, Zitong and Li, Xiaobai and Zhao, Guoying},
  booktitle={Proceedings of the IEEE/CVF Conference on Computer Vision and Pattern Recognition},
  pages={10631--10642},
  year={2021}
}

@article{smg2023ijcv,
  title={{SMG}: A Micro-gesture Dataset Towards Spontaneous Body Gestures for Emotional Stress State Analysis},
  author={Chen, Haoyu and Shi, Henglin and Liu, Xin and Li, Xiaobai and Zhao, Guoying},
  journal={International Journal of Computer Vision},
  volume={131},
  number={5},
  pages={1346--1366},
  year={2023},
  doi={10.1007/s11263-023-01761-6}
}

@inproceedings{pointtad2022,
  title={{PointTAD}: Multi-Label Temporal Action Detection with Learnable Query Points},
  author={Tan, Jing and Zhao, Xiaotong and Shi, Xintian and Kang, Bin and Wang, Limin},
  booktitle={Advances in Neural Information Processing Systems},
  volume={35},
  year={2022}
}

@inproceedings{guo2023miga_online,
  author={Guo, Xu Peng and Peng, Wei and Huang, Hexiang and Xia, Zhaoqiang},
  title={Micro-gesture Online Recognition with Graph-convolution and Multiscale Transformers for Long Sequence},
  booktitle={Proceedings of the IJCAI 2023 Workshop on Micro-gesture Analysis for Hidden Emotion Understanding (MiGA 2023)},
  series={CEUR Workshop Proceedings},
  volume={3522},
  year={2023}
}

@inproceedings{wang2024miga_dualstream,
  author={Wang, Yuhan and Linghu, Kerui and Huang, Hexiang and Xia, Zhaoqiang},
  title={Micro-gesture Online Recognition with Dual-stream Multi-scale Transformer in Long Videos},
  booktitle={Proceedings of the IJCAI 2024 Workshop on Micro-gesture Analysis for Hidden Emotion Understanding (MiGA 2024)},
  series={CEUR Workshop Proceedings},
  volume={3848},
  year={2024}
}

@inproceedings{liu2024miga_query,
  author={Liu, Pengyu and Wang, Fei and Li, Kun and Chen, Guoliang and Wei, Yanyan and Tang, Shengeng and Wu, Zhiliang and Guo, Dan},
  title={Micro-gesture Online Recognition using Learnable Query Points},
  booktitle={Proceedings of the IJCAI 2024 Workshop on Micro-gesture Analysis for Hidden Emotion Understanding (MiGA 2024)},
  series={CEUR Workshop Proceedings},
  volume={3848},
  year={2024}
}

@inproceedings{lin2017focal,
  title={Focal Loss for Dense Object Detection},
  author={Lin, Tsung-Yi and Goyal, Priya and Girshick, Ross and He, Kaiming and Doll{\'a}r, Piotr},
  booktitle={Proceedings of the IEEE International Conference on Computer Vision},
  pages={2980--2988},
  year={2017}
}

@inproceedings{zheng2020distanceiou,
  title={Distance-{IoU} Loss: Faster and Better Learning for Bounding Box Regression},
  author={Zheng, Zhaohui and Wang, Ping and Liu, Wei and Li, Jinze and Ye, Rongguang and Ren, Dongwei},
  booktitle={Proceedings of the AAAI Conference on Artificial Intelligence},
  volume={34},
  number={07},
  pages={12993--13000},
  year={2020}
}

@inproceedings{videomae2022,
  title={Video{MAE}: Masked Autoencoders Are Data-Efficient Learners for Self-Supervised Video Pre-Training},
  author={Tong, Zhan and Song, Yibing and Wang, Jue and Wang, Limin},
  booktitle={Advances in Neural Information Processing Systems},
  volume={35},
  pages={10078--10093},
  year={2022}
}

@article{calvo2010affect,
  title={Affect Detection: An Interdisciplinary Review of Models, Methods, and Their Applications},
  author={Calvo, Rafael A. and {D'Mello}, Sidney},
  journal={IEEE Transactions on Affective Computing},
  volume={1},
  number={1},
  pages={18--37},
  year={2010},
  doi={10.1109/T-AFFC.2010.1}
}

@article{li2026bench,
  title={{MA-Bench}: Towards Fine-grained Micro-Action Understanding},
  author={Li, Kun and Gu, Jihao and Wang, Fei and Wu, Zhiliang and Fan, Hehe and Guo, Dan},
  journal={arXiv preprint arXiv:2603.26586},
  year={2026}
}

@inproceedings{li2023data,
  title={Data Augmentation for Human Behavior Analysis in Multi-Person Conversations},
  author={Li, Kun and Guo, Dan and Chen, Guoliang and Liu, Feiyang and Wang, Meng},
  booktitle={Proceedings of the 31st ACM International Conference on Multimedia},
  pages={9516--9520},
  year={2023}
}

@article{li2023joint,
  title={Joint Skeletal and Semantic Embedding Loss for Micro-Gesture Classification},
  author={Li, Kun and Guo, Dan and Chen, Guoliang and Peng, Xinge and Wang, Meng},
  journal={arXiv preprint arXiv:2307.10624},
  year={2023}
}

@article{chen2024prototype,
  title={Prototype Learning for Micro-Gesture Classification},
  author={Chen, Guoliang and Wang, Fei and Li, Kun and Wu, Zhiliang and Fan, Hehe and Yang, Yi and Wang, Meng and Guo, Dan},
  journal={arXiv preprint arXiv:2408.03097},
  year={2024}
}

@article{gu2025mm,
  title={{MM-Gesture}: Towards Precise Micro-Gesture Recognition through Multimodal Fusion},
  author={Gu, Jihao and Wang, Fei and Li, Kun and Wei, Yanyan and Wu, Zhiliang and Guo, Dan},
  journal={arXiv preprint arXiv:2507.08344},
  year={2025}
}

@inproceedings{gu2025motion,
  title={Motion Matters: Motion-Guided Modulation Network for Skeleton-Based Micro-Action Recognition},
  author={Gu, Jihao and Li, Kun and Wang, Fei and Wei, Yanyan and Wu, Zhiliang and Fan, Hehe and Wang, Meng},
  booktitle={Proceedings of the 33rd ACM International Conference on Multimedia},
  pages={5461--5470},
  year={2025}
}

@inproceedings{li2025prototypical,
  title={Prototypical Calibrating Ambiguous Samples for Micro-Action Recognition},
  author={Li, Kun and Guo, Dan and Chen, Guoliang and Fan, Chunxiao and Xu, Jingyuan and Wu, Zhiliang and Fan, Hehe and Wang, Meng},
  booktitle={Proceedings of the AAAI Conference on Artificial Intelligence},
  volume={39},
  number={5},
  pages={4815--4823},
  year={2025}
}

@article{liu2025online,
  title={Online Micro-Gesture Recognition Using Data Augmentation and Spatial-Temporal Attention},
  author={Liu, Pengyu and Li, Kun and Wang, Fei and Wei, Yanyan and She, Junhui and Guo, Dan},
  journal={arXiv preprint arXiv:2507.09512},
  year={2025}
}

@inproceedings{thumos14,
  title={{THUMOS} Challenge: Action Recognition with a Large Number of Classes},
  author={Idrees, Haroon and Zamir, Amir R. and Jiang, Yu-Gang and Gorban, Alex and Laptev, Ivan and Sukthankar, Rahul and Shah, Mubarak},
  booktitle={ECCV Workshop on Action Recognition with a Large Number of Classes},
  year={2014}
}

@inproceedings{activitynet2015,
  title={{ActivityNet}: A Large-Scale Video Benchmark for Human Activity Understanding},
  author={Caba Heilbron, Fabian and Escorcia, Victor and Ghanem, Bernard and Carlos Niebles, Juan},
  booktitle={Proceedings of the IEEE/CVF Conference on Computer Vision and Pattern Recognition},
  pages={961--970},
  year={2015}
}

@article{liu2025survey,
  title={A survey on fmri-based brain decoding for reconstructing multimodal stimuli},
  author={Liu, Pengyu and Dong, Guohua and Guo, Dan and Li, Kun and Li, Fengling and Yang, Xun and Wang, Meng and Ying, Xiaomin},
  journal={arXiv preprint arXiv:2503.15978},
  year={2025}
}

@article{Noroozietal2021,
  author  = {Noroozi, Fatemeh and Corneanu, Ciprian Adrian and Kaminska, Dorota and Sapinski, Tomasz and Escalera, Sergio and Anbarjafari, Gholamreza},
  title   = {Survey on Emotional Body Gesture Recognition},
  journal = {IEEE Transactions on Affective Computing},
  year    = {2021},
  volume  = {12},
  pages   = {505--523},
  doi     = {10.1109/taffc.2018.2874986}
}

@article{Poppeetal2024,
  author  = {Poppe, Ronald and van der Zee, Sophie and Taylor, Paul J. and Anderson, Ross J. and Veltkamp, Remco C.},
  title   = {Mining Bodily Cues to Deception},
  journal = {Journal of Nonverbal Behavior},
  year    = {2024},
  volume  = {48},
  pages   = {137--159},
  doi     = {10.1007/s10919-023-00450-9}
}

@article{Ekman1969,
  author  = {Ekman, Paul and Friesen, Wallace V.},
  title   = {Nonverbal leakage and clues to deception},
  journal = {PsycEXTRA Dataset},
  year    = {1969},
  doi     = {10.1037/e525532009-012}
}

@inproceedings{loshchilov2017decoupled,
  title={Decoupled Weight Decay Regularization},
  author={Loshchilov, Ilya and Hutter, Frank},
  booktitle={International Conference on Learning Representations},
  year={2019}
}

@article{shen2026spatial,
  title={Spatial-Temporal Decoupled Adapter for Micro-gesture Online Recognition},
  author={Shen, Xucheng and Li, Kun and Wang, Fei and Qian, Wei and Jiang, Jin and Guo, Dan},
  journal={arXiv preprint arXiv:2606.07355},
  year={2026}
}

@inproceedings{shang2025cross,
  title={Cross-modal Feature Enhancement and Contrastive Alignment for Micro-gesture Recognition},
  author={Shang, Tuyun and Hao, Yanbin and Pei, Ming and Li, Kun and Ben, Huixia and Wang, Shuo},
  booktitle={Chinese Conference on Pattern Recognition and Computer Vision (PRCV)},
  pages={203--217},
  year={2025},
  organization={Springer}
}

@inproceedings{hao2022group,
  title={Group contextualization for video recognition},
  author={Hao, Yanbin and Zhang, Hao and Ngo, Chong-Wah and He, Xiangnan},
  booktitle={Proceedings of the ieee/cvf conference on computer vision and pattern recognition},
  pages={928--938},
  year={2022}
}

@article{hao2022attention,
  title={Attention in attention: Modeling context correlation for efficient video classification},
  author={Hao, Yanbin and Wang, Shuo and Cao, Pei and Gao, Xinjian and Xu, Tong and Wu, Jinmeng and He, Xiangnan},
  journal={IEEE Transactions on Circuits and Systems for Video Technology},
  volume={32},
  number={10},
  pages={7120--7132},
  year={2022},
  publisher={IEEE}
}

@inproceedings{meng2025online,
  title={Online micro-gesture recognition in long videos via spatiotemporal feature encoding and query-based temporal detection},
  author={Meng, Chutian and Ma, F. and Zhang, C. and Miao, J. and Yang, Y. and Zhuang, Y.},
  booktitle={Proceedings of the IJCAI 2025 Workshop on Micro-gesture Analysis for Hidden Emotion Understanding (MiGA 2025)},
  year={2025}
}

\end{document}